\documentclass[letterpaper,10pt,conference]{ieeeconf}
\IEEEoverridecommandlockouts
\usepackage{iftex}
\ifXeTeX
\usepackage{fontspec}
\IfFontExistsTF{Times New Roman}{\setmainfont{Times New Roman}}{\setmainfont{TeX Gyre Termes}}
\fi
\usepackage{cite}
\usepackage{amsmath,amssymb}
\usepackage{graphicx}
\usepackage{booktabs}
\usepackage{array}
\usepackage{siunitx}
\usepackage{url}

\graphicspath{{figures/}}

\newif\ifanonymous
\anonymousfalse

\title{Embodied Snap: Octopus-Inspired Distributed Reach-and-Attach\\
with a Speed-Limited Soft Arm \vspace{-4mm}}

\ifanonymous
\author{Anonymous Authors}
\else
\author{Linxin Hou, Zhihang Qin, Heyang Zou, Qirui Wu, Peiyi Wang, Muhammad Sunny Nazeer$^{*}$, \\ Yongxin Guo$^{*}$, and Cecilia Laschi$^{*}$
\thanks{This work was supported by the NUS bridging fund (AI-Driven Soft Robots for Marine and Unstructured Environments).}%
\thanks{L. Hou is with the Department of Electrical and Computer Engineering, National University of Singapore, Singapore. (Email: hou.linxin@u.nus.edu)}%
\thanks{Z. Qin, H. Zou, Peiyi Wang, M. S. Nazeer, and C. Laschi are with the Department of Mechanical Engineering, National University of Singapore, Singapore.}%
\thanks{Q. Wu is with the Department of Computer Science, National University
of Singapore, Singapore.}%
\thanks{Y. Guo is with the Department of Electrical and Computer Engineering, National University of Singapore. He is also with the Department of Electrical Engineering, City University of Hong Kong, Hong Kong SAR, China.}
\thanks{C. Laschi is also with the Advanced Robotic Centre, National University of Singapore, Singapore.
}
}
\fi

\begin{document}
\def\IEEEtitletopspaceextra{0.2in}
\IEEEaftertitletext{\vspace{-1.2\baselineskip}}
\maketitle
\thispagestyle{empty}
\pagestyle{empty}

\begin{abstract}
Reach-and-attach of soft robotic arms with passive suction requires accurate targeting and sufficient contact speed, yet geared actuators can impose a speed limit that improved trajectory tracking alone cannot overcome. This paper proposes an embodied snap controller that separates slow servo-driven preloading from rapid elastic release, enabling a compliant arm to move beyond its direct tendon-driven speed limit. Octopus biology motivates the controller’s section-wise organizational prior, rather than reproduction of the octopus nervous system. A learned policy shared across three sections selects preloads, aim, tendon slack, and release timing, determining where, how, and when to load and release the body. The policy is optimized offline using a hardware-validated recurrent model within experimentally supported bounds. Across five optimization seeds and 400 unseen simulated targets, attachment success is $(73\pm4)\%$ at a \SI{5}{cm} lateral tolerance, and the shared policy reaches the matched centralized controller's mean final reward after a median \SI{17}{\percent} of the common evaluation budget. Hardware characterization achieves tip speeds of \SIrange{1.56}{1.64}{m/s}, at least \SI{108}{\percent} above direct tendon-driven release. In 18 open-loop hardware trials across six placements, 17 exceed the \SI{1}{m/s} snap threshold and nine retrieve the object, with successful retrieval at five placements. These results demonstrate a practical division of responsibility in the control problem: learned control prepares the body, and passive body mechanics execute the rapid movement needed for dynamic reach-and-attach.
\end{abstract}

\vspace{-5mm}

\section{Introduction}
\begin{figure*}[t]
\centering
\begin{minipage}[c]{0.3\textwidth}
\centering
\includegraphics[width=\linewidth]{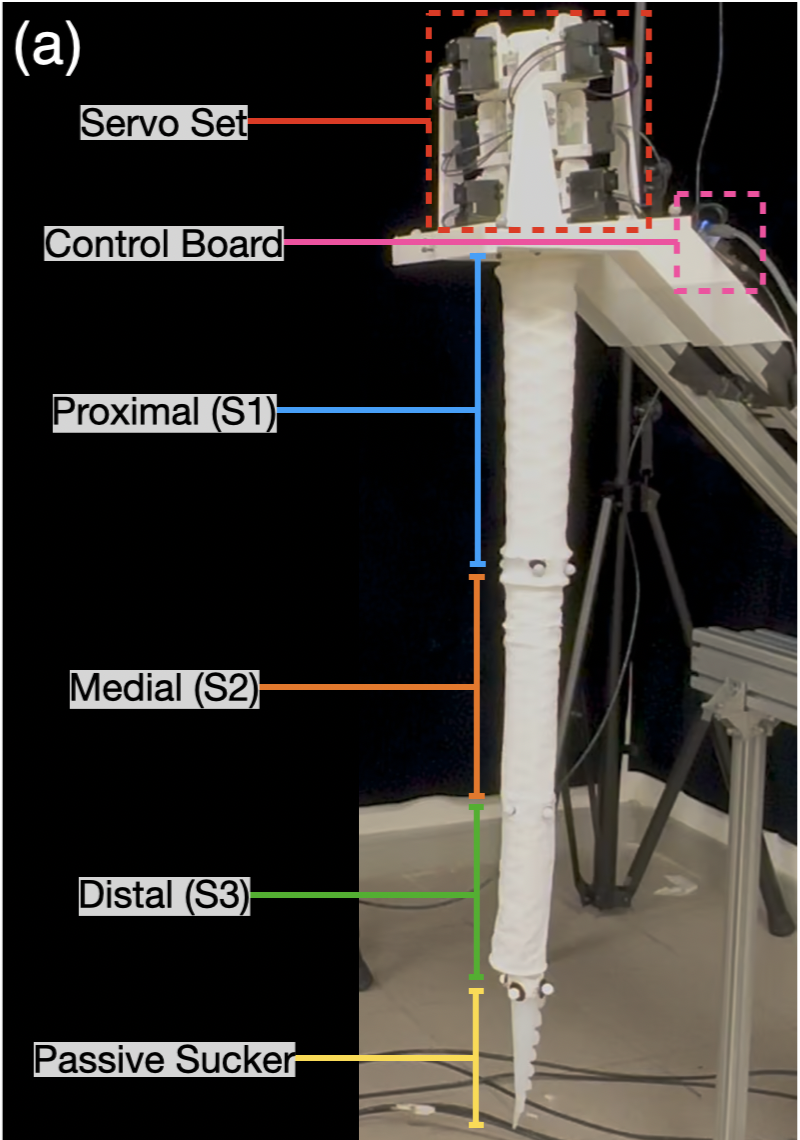}
\end{minipage}\hfill
\begin{minipage}[c]{0.67\textwidth}
\centering
\includegraphics[width=\linewidth]{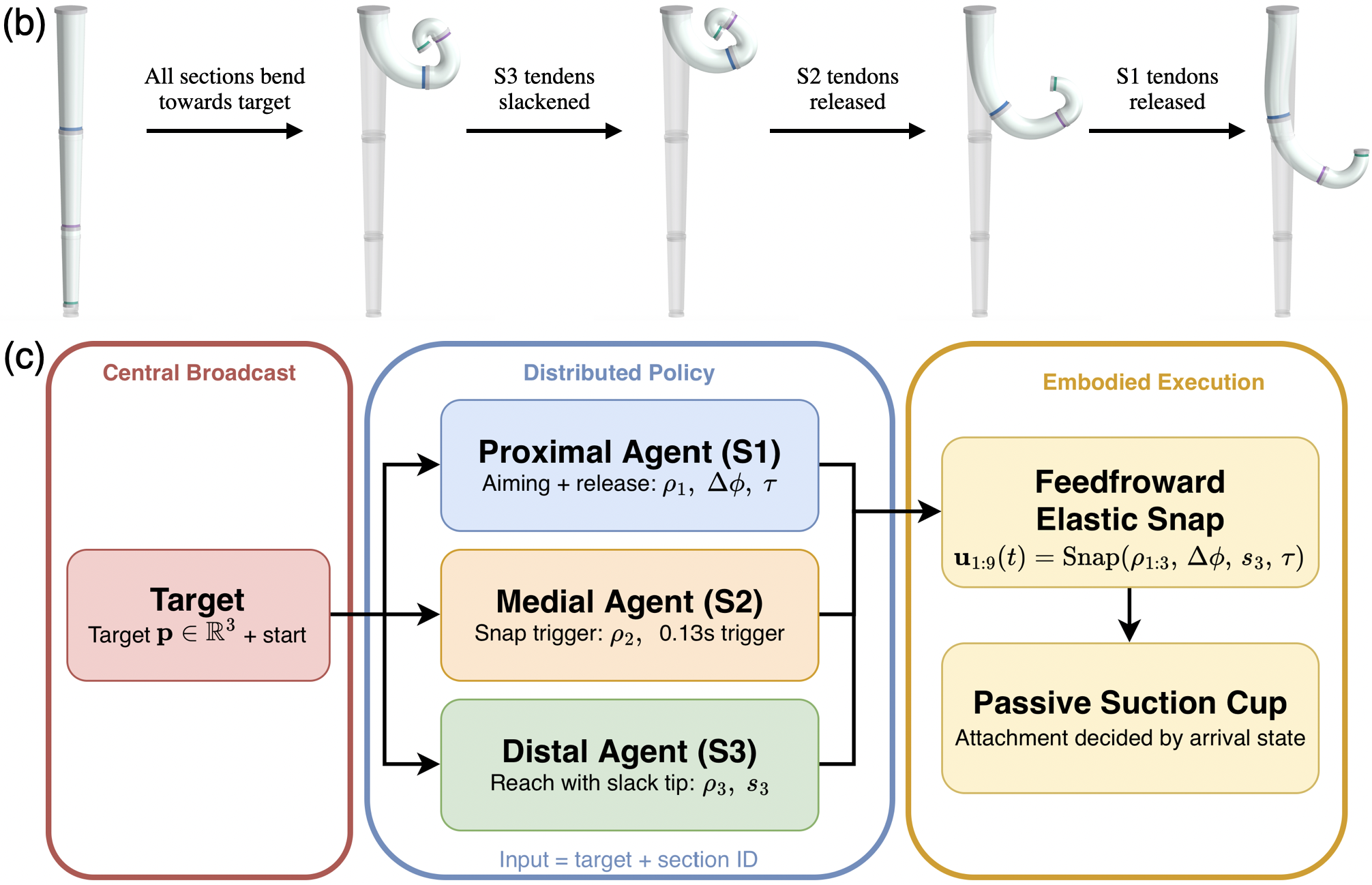}
\end{minipage}
\caption{Embodied snap-control system. (a) Three-section soft arm. (b) Simulated primitive: preload toward target, then release S3, S2, and S1 sequentially. (c) A target-conditioned shared policy assigns section-specific preload, aim, slack, and release commands for a feedforward snap and passive attachment.}
\label{fig:system}
\vspace{-4mm}
\end{figure*}

Embodied intelligence emphasizes how morphology, material response, and environmental interaction contribute to robotic behavior alongside control \cite{pfeifer2007self,sitti2021physical,mengaldo2022concise,milana2025physical}. In soft robots, compliance can store energy and shape motion, providing a physical resource for task execution. Exploiting this resource requires the controller to establish conditions under which the body's response produces a useful action. This paper investigates this coordination of control and mechanics through dynamic reach-and-attach with a three-section tendon-driven soft arm with passive suction cups (Fig.~\ref{fig:system}(a)).

Attachment requires sufficient target overlap and contact speed to compress the cups and establish a seal. On our platform, geared servos provide controlled quasi-static positioning, but direct tendon-driven release remains below the speeds associated with attachment in hardware tests. Improving positioning accuracy alone therefore does not address the measured speed limitation. Elastic energy storage offers a way to separate slow loading from rapid motion \cite{ilton2018principles,pal2020elastic,gorissen2020inflatable,feng2025impulsive}. Therefore, we use the servos to preload the arm, then sequentially release the tendons to generate a rapid elastic snap toward the target (Fig.~\ref{fig:system}(b)).

A fast release does not ensure target contact. The preloaded configuration and release sequence jointly determine the subsequent motion, while hysteresis and tendon coupling complicate prediction. The control problem is to select preload, aim, slack, and release timing that coordinate speed and spatial approach. We address this problem with a recurrent simulator validated against hardware measurements and a compact policy optimized within hardware-supported bounds. Octopus biology motivates an organizational prior to section-wise control, rather than an attempt to reproduce the octopus nervous system \cite{sumbre2001control,hochner2023embodied}. We implement this prior using a shared policy evaluated by each section, conditioned on the common target and section identity, to select its primitive parameters (Fig.~\ref{fig:system}(c)). The nominal snap proceeds without state feedback or inter-section messages. The learned control prepares the target-dependent response, while passive body mechanics generate the rapid deformation. The biological inspiration concerns the control organization, while elastic release supplies the speed amplification.

The paper makes three contributions:
\begin{enumerate}
\setlength{\itemsep}{0pt}
\setlength{\parsep}{0pt}
\setlength{\topsep}{2pt}
\item a hardware-characterized elastic snap that increases the measured soft-arm tip speed by at least 108\% relative to direct tendon-driven release under the tested operating conditions;
\item a hardware-gated recurrent simulator and a 155-parameter shared policy for selecting preload, aim, slack, and release;
\item matched controller comparisons, a release-rule ablation, and a hardware demonstration of 18-trials.
\end{enumerate}
The source code for this paper is available at \url{https://anonymous.4open.science/r/EmbodiedSnap/}.

\section{Related Work}
Elastic energy storage enables rapid motion through different physical mechanisms. Ilton et al.\ analyze coupled dynamics of motors, springs, and latches \cite{ilton2018principles}, while Pal et al.\ use prestressed elastomers for rapid recovery \cite{pal2020elastic}. Gorissen et al.\ convert slow inflation into jumping through shell snap-through \cite{gorissen2020inflatable}, an instability-based form of impulsive actuation \cite{feng2025impulsive}. Our arm uses tendon-mediated preloading and sequential release, with loading and release conditions selected for target-directed motion. Here, elastic snap denotes this rapid release motion rather than an established snap-through instability.

Peripheral control of the octopus has inspired continuum robots and section-wise soft-arm architectures \cite{sumbre2001control,hochner2023embodied,mazzolai2019octopus,liu2026underwater,hou2026quantitative}. Modular policies also provide a reusable local structure in articulated robots \cite{wang2018nervenet,gupta2022metamorph}. SoftGM uses graph-attention messages to integrate proprioception and contact information across sections \cite{hou2026octopus}. Our shared policy selects section-specific parameters from the target and section identity, without exchanging state during nominal execution. Comparisons with centralized and unshared policies assess this parameterization.

Octopus suckers attach through compliant local deformation and a pressure seal \cite{kier2002structure}, and suction cups have been integrated with octopus-inspired continuum arms for grasping \cite{margheri2012soft,mazzolai2019octopus}. Our passive cups provide the terminal attachment mechanism following target-directed motion. Hardware retrieval is evaluated separately from simulated geometric and speed criteria, which do not model suction sealing.

Learned forward models support controller training, with transfer depending on model fidelity and data coverage \cite{falotico2025learning}. Berdica et al.\ use recurrent models to train feedback policies \cite{berdica2024reinforcement}, while Nazeer et al.\ address training-to-reality mismatch through adaptive compensation and coaching \cite{TRO_nazeer2024}. Tang et al.\ combine offline learning with online adaptation \cite{tang2026general}. For dynamic manipulation, Haggerty et al.\ use Koopman models for inertial feedback control \cite{haggerty2023inertial}, while SofToss learns an open-loop throwing policy from an actuation-to-landing-position model \cite{bianchi2024softoss}. Our recurrent model predicts post-release trajectories, allowing spatial approach and speed to be evaluated jointly. Policy search is restricted to hardware-supported loading and timing ranges and penalizes weak data support and model-ensemble disagreement. The resulting policy executes without online adaptation or feedback during the snap.

\begin{figure*}[t]
\centering
\includegraphics[width=0.88\textwidth]{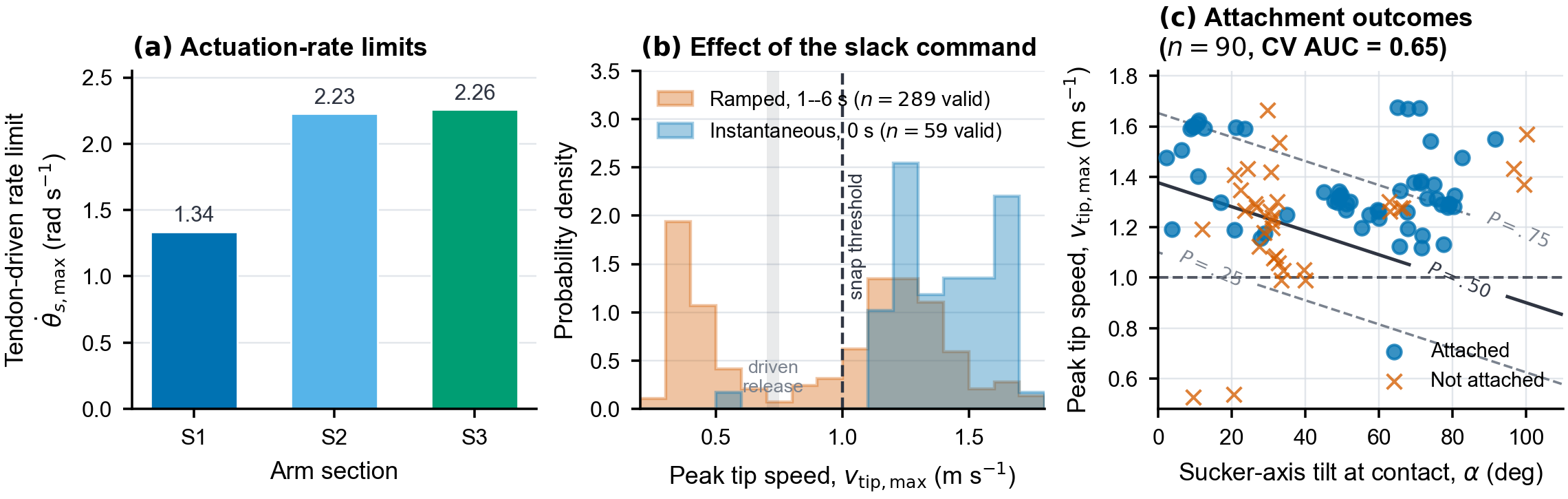}
\vspace{-3mm}
\caption{Actuation constraint and attachment evidence. (a) Tendon-driven angular-rate ceilings from calibrated tendon kinematics and the rated 12-V no-load servo speed. (b) Tip-speed distributions for ramped and instantaneous distal release. The shaded interval is the direct-release range and the dashed line is the empirical \SI{1}{m/s} attachment threshold. (c) Outcomes from 90 target trials over tip speed and cup-axis tilt. Contours show the fitted two-variable model.}
\label{fig:limits}
\vspace{-3mm}
\end{figure*}

\begin{figure*}[t]
\centering
\includegraphics[width=0.88\textwidth]{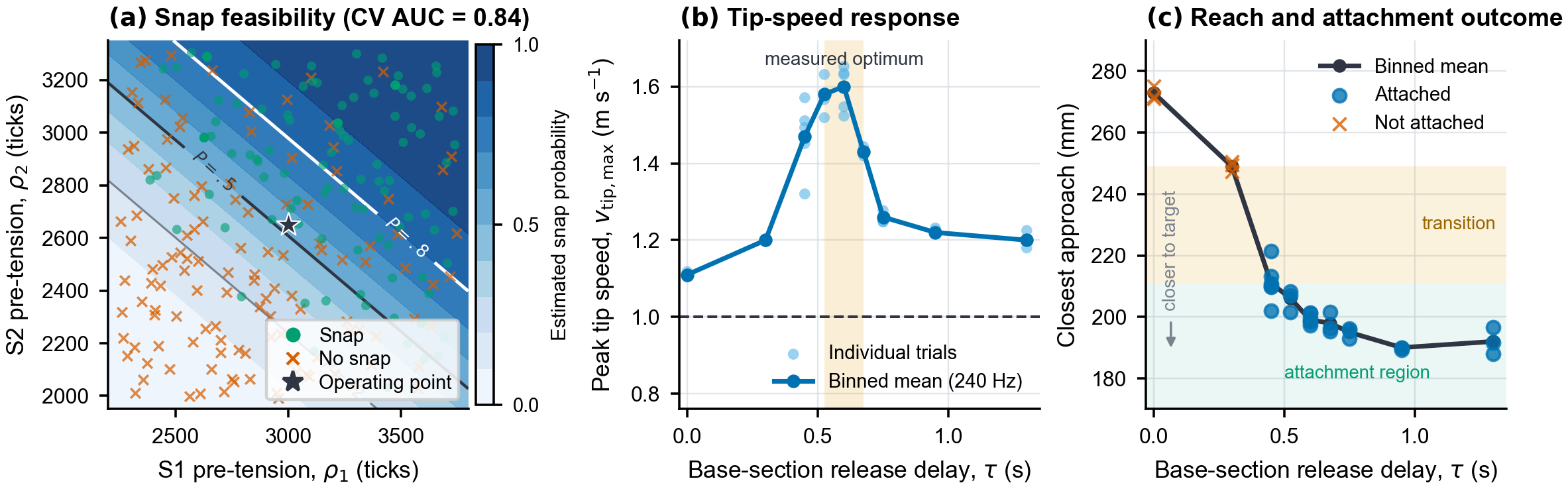}
\vspace{-3mm}
\caption{Hardware snap preconditions and timing. (a) Estimated probability of exceeding \SI{1}{m/s} over proximal preloads; the star is the chosen operating point. (b) Peak tip speed over S1 release delay. (c) Closest approach and attachment outcome in the same sweep. Shaded regions mark the empirically supported operating intervals.}
\label{fig:snap}
\vspace{-3mm}
\end{figure*}

\section{System and Embodied Snap}
\subsection{Platform and actuator limit}
The soft arm platform used in this paper is shown in Fig.~\ref{fig:system}(a). The vertically suspended arm has three tapered helical-lattice sections of lengths \SIlist{324;230;176}{mm} from the base to the tip (S1--S3). The arm is printed from TPU 95A using a Bambu Lab H2D printer, with rigid PLA connectors and mounting components. Three tendons per section are driven by Robotis Dynamixel XM430-W350-R servos with non-backdrivable gearboxes. Distal tendons traverse the proximal sections, coupling their load. The arm has a \SI{150}{mm} octopus-like compliant end effector made by Ecoflex GEL 2 that carries multiple passive suction cups. Four arm bodies and the target are tracked at \SI{240}{Hz}. The controller and learned-model states use a \SI{50}{Hz} grid.

The servo logs give \SI{15.8}{\micro\metre} of tendon travel per tick (interquartile range (IQR) \SIrange{14.4}{18.0}{\micro\metre}). For a taut tendon $j$ terminating in section $s(j)$, the calibrated displacement satisfies
\begin{equation}
k w_j=\sum_{p=1}^{s(j)}\left[
d_p\theta_p\cos(\phi_p-\alpha_j)+\Delta L_p\right],
\label{eq:tendon}
\end{equation}
where $w_j$ is servo rotation in ticks, $k$ is tendon travel per tick, $d_p$ is the effective arm lever, $\theta_p$ and $\phi_p$ are section bend magnitude and direction, $\alpha_j$ is tendon azimuth, and $\Delta L_p$ is axial shortening. The sum represents the distal tendons that traverse the proximal sections. A slack tendon gives an inequality rather than equality, so the parameters were fitted to the taut envelope by quantile regression. The resulting ratios are $d_p/k=\SIlist{2351;1411;1390}{ticks/rad}$.

During commanded release, measured rates reach \SIrange{100}{107}{\percent} of the rated no-load value. A first-order fit to 291 bursts gives a \SI{3000}{tick/s} maximum and \SI{0.03}{s} time constant, consistent with the rated ceiling $\dot w_{\max}=\SI{3140}{tick/s}$. The rated value is used in Eq.~\eqref{eq:rate-limit} as a conservative upper bound. For a pure section bend, Eq.~\eqref{eq:tendon} gives
\begin{equation}
\dot\theta_{p,\max}=\frac{\dot w_{\max}}{d_p/k}.
\label{eq:rate-limit}
\end{equation}
Hence, the S1--S3 tendon-driven angular-rate ceilings are \SIlist{1.34;2.23;2.26}{rad/s} (Fig.~\ref{fig:limits}(a)). Direct release produces only \SIrange{0.70}{0.75}{m/s} of tip speed (Fig.~\ref{fig:limits}(b)). In 90 labeled target trials, no attachment occurs below \SI{1}{m/s}, identifying a terminal speed condition that direct release does not meet.

\subsection{Embodied snap primitive}
The four-stage primitive (Fig.~\ref{fig:system}(b)) begins by aiming the arm and establishing section preloads $\rho_{1:3}$. The S3 tendons then slackened, and the folded lattice retains a curled tip as a mechanical latch. S2 is released over \SI{0.13}{s}, followed by S1 after delay $\tau$, producing a ballistic snap toward the target and a possible passive attachment. The snap arises from elastic recoil of the preloaded TPU lattice as tendon slackening releases its constraints, allowing stored strain energy to drive motion beyond the direct tendon-driven speed limit. The proposed policy selects preloads, tip slack, aim correction, and delay. Instantaneous distal slackening causes 98\% of 59 trials to exceed \SI{1}{m/s}. Ramping the same command over \SIrange{1}{6}{s} instead produces a slow servo-dragged release or a fast snap (Fig.~\ref{fig:limits}(b)). A grouped cross-validated classifier of the speed-threshold event from proximal preload reaches an area under curve (AUC) of 0.84, only 27\% of the explored parameter box has predicted event probability higher than 0.8 (Fig.~\ref{fig:snap}(a)). At the selected operating point, tip speed reaches \SIrange{1.56}{1.64}{m/s}, and section rates are \SIlist{412;458;866}{\degree/s}. These are respectively $5.4$, $3.6$, and $6.7$ times the tendon-driven ceilings in Eq.~\eqref{eq:rate-limit}.

In an independent 33-trial sweep, the peak speed is maximized at $\tau=\text{\SIrange{0.525}{0.60}{s}}$, while the closest approach improves to \SI{1.3}{s} (Fig.~\ref{fig:snap}(b,c)). Delays $\tau\leq\SI{0.30}{s}$ yield 0/6 attachments, versus 27/27 for $\tau\geq\SI{0.45}{s}$. Therefore, policy timing is restricted to the tested \SIrange{0.45}{0.675}{s} interval. Representative mistimed failures can be referred to in the accompanying video. Among 42 trials that arrive within \SI{150}{mm} laterally at the target height and above \SI{1}{m/s}, 93--95\% attach despite the large observed tilts of the cup-axis. A descriptive speed-angle logistic model (Fig.~\ref{fig:limits}(c)) achieves a grouped cross-validated AUC 0.65 under nonuniform sampling. We therefore treat \SI{1}{m/s} as a necessary rig-specific condition and require block retention for hardware success.

\section{Hardware-Gated Model and Policy}

\subsection{Hardware data and dynamics-model validation}
A gated recurrent unit (GRU) predicts post-release motion from recent body states and servo commands, enabling offline policy evaluation. The dataset contains 538 independent hardware acquisition groups: 203 free-space trials, 76 target-conditioned trials, 20 sealed tests, 28 pilot groups, and 211 characterization groups. The 440/78/20 training/validation/sealed-test split keeps overlapping windows within their acquisition groups. The independent 33-trial delay sweep is excluded from the fitting (Table~\ref{tab:evidence}).

Let $\mathbf x_t\in\mathbb R^{21}$ contain three tracked body positions, the tip axis, and nine servo positions, and let $\mathbf u_t\in\mathbb R^9$ contain the servo goals. The selected two-layer, width-96 GRU predicts a residual over a constant-velocity prior:
\begin{equation}
\hat{\mathbf x}_{t+1}=f_{\mathrm{cv}}(\mathbf x_t,\mathbf x_{t-1})+
F_{\boldsymbol\psi}(\mathbf x_{t-19:t},\mathbf u_{t-19:t+5}).
\label{eq:dynamics}
\end{equation}
Here $\hat{\mathbf x}_{t+1}$ is the predicted next state, $f_{\mathrm{cv}}$ is the constant-velocity prior, $F_{\boldsymbol\psi}$ is the GRU residual model with parameters $\boldsymbol\psi$, and the indexed sequences provide a history \SI{0.4}{s} and \SI{0.1}{s} command look-ahead at \SI{50}{Hz}. During an $H=25$ step rollout, each predicted state is inserted into the next history:
\begin{equation}
\begin{aligned}
\hat{\mathbf x}_{t+h+1}
&=f_{\mathrm{cv}}(\hat{\mathbf x}_{t+h},\hat{\mathbf x}_{t+h-1})\\
&\quad+F_{\boldsymbol\psi}(\hat{\mathcal H}_{t+h},\mathcal U_{t+h}),
\end{aligned}
\label{eq:rollout}
\end{equation}
where $h$ is the rollout index, $\hat{\mathcal H}_{t+h}$ is the updated state-history window, and $\mathcal U_{t+h}$ is the corresponding command window. For $h=0,\ldots,H-1$, no measured state is injected after initialization. The model is trained to reduce the normalized mean squared prediction error over 25 steps, with step weights proportional to $h^{-1/2}$. Errors from snap and fast-motion windows receive factors of 5 and 3 respectively. Three random seeds are used together to optimize the policy.

\begin{figure*}[t]
\centering
\includegraphics[width=0.88\textwidth]{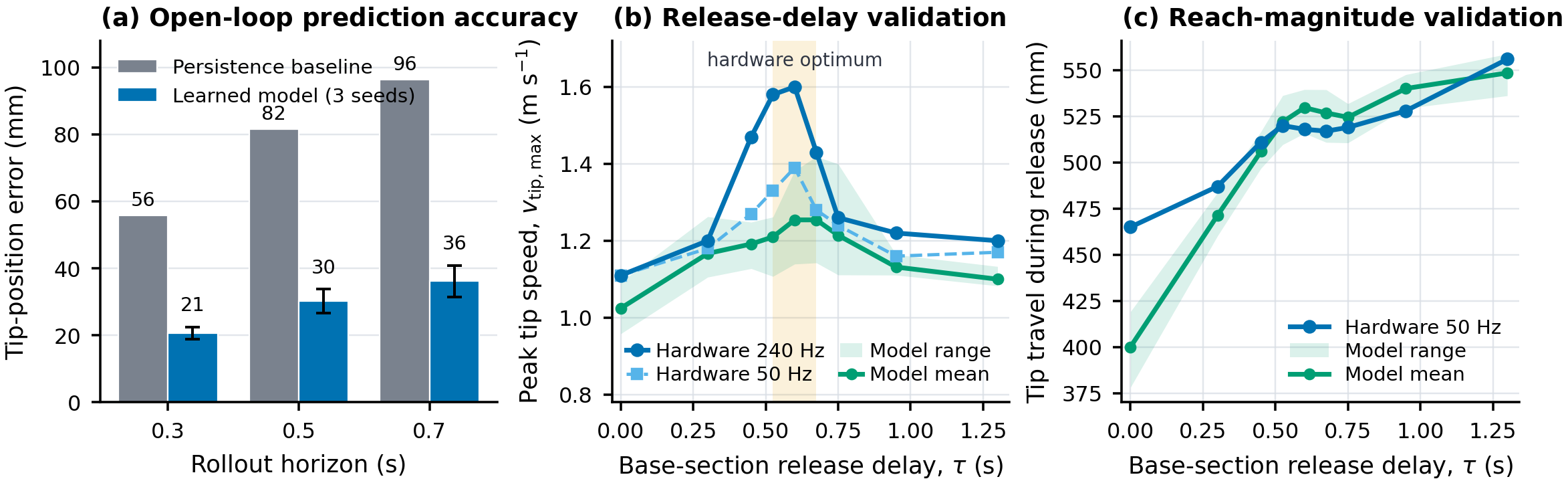}
\vspace{-3mm}
\caption{Hardware gates for controller development. (a) Validation-set open-loop tip error for persistence and the GRU. (b) Measured and predicted peak speed over the excluded delay sweep. (c) Measured and predicted tip travel in the same sweep, bands in (b,c) span model seeds.}
\label{fig:gates}
\vspace{-6mm}
\end{figure*}

Figure~\ref{fig:gates}(a) reports validation tip errors of \SIlist{21;30;36}{mm} at \SIlist{0.3;0.5;0.7}{s}, versus \SIlist{56;82;96}{mm} for persistence. After model selection, a single sealed evaluation over 11,284 windows gives seed ranges of \SIrange{22.3}{23.1}{mm}, \SIrange{31.8}{33.6}{mm}, and \SIrange{37.3}{39.5}{mm}, versus \SIlist{72;106;125}{mm} for persistence. These GRU errors span 3.1--5.4\% of the undeformed arm length, excluding the end effector. Each model seed places the speed optimum within the measured attachment plateau, although the maximum speed is underestimated (Fig.~\ref{fig:gates}(b)), predicted travel differs from the hardware measurements by 3--4\% (Fig.~\ref{fig:gates}(c)). These checks support policy search within the tested region.

The GRU models only the rapid motion after release. The preceding several-second preload is represented by a quasi-static hold model, whose terminal configuration initializes the GRU rollout. A variable-strain Cosserat model \cite{renda2018discrete} describes the quasi-static raise and pass-through tendon routing. The terminal hold used to initialize Eq.~\eqref{eq:rollout} is a distance-weighted average of the 12 nearest recorded holds in $(\rho_1,\rho_2,\rho_3,s_3)$, scaled by $(1500,1500,1500,1000)$ ticks and weighted by inverse distance with a 0.01 floor. The support confidence $C_h=0.5+0.5\exp[-(\bar\delta/0.3)^2]$ decreases commands with a large mean neighbor distance $\bar\delta$. After slackening, the model-predicted pre-release configuration with a median tip error of \SI{59.6}{mm} (\SI{8.2}{\percent} of $L_{\mathrm{arm}}$), an axis error of \SI{13.7}{\degree}, reward correlation 0.89 with recorded hold replay, and attachment AUC 0.68 versus 0.72 from recorded holds. These errors motivate both uncertainty penalties during optimization and the full geometric tolerance sweep.

\subsection{Shared, target-conditioned policy}
The controller maps a target position to six parameters of the
preload-and-release primitive: three section preloads $(\rho_1,\rho_2,\rho_3)$, distal slack $s_3$, an aiming correction $\Delta\phi$, and the $S_1$ release delay $\tau$. Let $\mathbf b$ be the arm base, $\mathbf n$ its measured free-hanging axis, and $\beta(\mathbf p)$ the bearing of target $\mathbf p$ about $\mathbf n$. Because the dynamics model is validated around $\phi_0=\SI{210}{\degree}$, a target is first rotated to the canonical bearing $\phi'_0=\phi_0+\Delta_{\mathrm{str}}=\SI{220}{\degree}$:
\begin{equation}
\tilde{\mathbf p}=\mathbf b+
\mathbf R_{\mathbf n}\!\left(\phi'_0-\beta(\mathbf p)\right)
(\mathbf p-\mathbf b),
\label{eq:canonical-target}
\end{equation}
where $\Delta_{\mathrm{str}}=\SI{10}{\degree}$ is the measured offset between commanded and strike bearing. The simulated command is $\phi_0+\Delta\phi$ and the hardware command is $\beta-\Delta_{\mathrm{str}}+\Delta\phi$. This reduction is supported by the \SI{120}{\degree} tendon relabelling and by the measured azimuth invariance within the recorded sector.

For the canonical target, let $\tilde{\mathbf q}=\tilde{\mathbf p}-\mathbf b$ and $\mathbf g=(d,e,r)=(\|\tilde{\mathbf q}\|_2,\tilde q_y, \sqrt{\tilde q_x^2+\tilde q_z^2})$, where $d$ is base-to-target distance, $e$ is elevation, and $r$ is horizontal radial distance. Section $i$ receives $\mathbf o_i=[(\mathbf g-\boldsymbol\mu)\oslash\boldsymbol\sigma_o,\mathbf e_i] \in\mathbb R^6$, where $\mathbf e_i$ is its one-hot identity and the normalization statistics are computed from the training targets. All sections evaluate the same network:
\begin{align}
\mathbf h_i^{(1)}&=\tanh(\mathbf W_1\mathbf o_i+\mathbf b_1),\notag\\
\mathbf h_i^{(2)}&=\tanh(\mathbf W_2\mathbf h_i^{(1)}+\mathbf b_2),\notag\\
\mathbf u_i&=\tanh(\mathbf W_3\mathbf h_i^{(2)}+\mathbf b_3)
\in[-1,1]^3.
\label{eq:shared-policy}
\end{align}
Here $\mathbf W_1\in\mathbb R^{8\times6}$, $\mathbf W_2\in\mathbb R^{8\times8}$, and $\mathbf W_3\in\mathbb R^{3\times8}$, giving $48+8+64+8+24+3=155$ shared parameters.

The network returns three dimensionless outputs per section. Each output that is assigned to an action $a_j\in[a_j^-,a_j^+]$ is decoded in its physical range as
\begin{equation}
a_j=a_j^-+(a_j^+-a_j^-)
\begin{cases}
(u_{ij}+1)/2, & a_j\ne s_3,\\
[(u_{ij}+1)/2]^2, & a_j=s_3,
\end{cases}
\label{eq:action-map}
\end{equation}
where the quadratic slack map provides greater resolution near zero. Here $\rho_i$ is the tendon take-up, or preload, of section $i$ in servo ticks. The assignments are $(u_{11},u_{12},u_{13})\mapsto(\rho_1,\Delta\phi,\tau)$, $u_{21}\mapsto\rho_2$, and $(u_{31},u_{32})\mapsto(\rho_3,s_3)$, with unused outputs discarded. Their ranges are $\rho_1\in[2400,3700]$ ticks, $\Delta\phi\in[-25,25]^{\circ}$, $\tau\in[0.45,0.675]$ s, $\rho_2\in[2100,3300]$ ticks, $\rho_3\in[2200,3600]$ ticks, and $s_3\in[0,1800]$ ticks. Each preload is also constrained by the measured directional workspace, $\rho_i\leftarrow\min\{\rho_i,0.9\rho_i^{\mathrm{max}}\}$, in the commanded bearing.

For the nominal feedforward policy, the S1 release time is
\begin{equation}
t_{\mathrm{rel}}=\Pi_{\mathcal T}(\tau),\qquad
\mathcal T=[0.45,0.675]\ \mathrm s,
\label{eq:release-time}
\end{equation}
where $\Pi_{\mathcal T}$ projects onto the hardware-validated interval. The target and start signals are broadcast before motion, and no state or inter-section message is used during the nominal snap. Covariance matrix adaptation evolution strategy (CMA-ES) \cite{hansen2001completely} is used with 120 generations, 24 candidates, and 48 targets per candidate ($1.38\times10^5$ simulator evaluations). Selection uses 64 held-out targets and all generalization results use 400 additional targets from a disjoint seed.

The baselines are a parameter-matched centralized controller (158 weights), a width-16 centralized controller (438 weights), and three unshared local networks (366 weights). All architectures use the same model ensemble, reward, constraints, targets, optimizer, budget, and seeds. The shared and unshared policies receive three broadcast target features, whereas the centralized controller outputs all six section-specific commands. For comparison with the learned target-only delay $\tau$, four alternative release rules keep the learned loading and aim fixed. A fixed-clock rule is released at \SI{0.50}{s} and the local $S_1$ triggers fires when its bending rate crosses \SI{15}{\degree/s}; the distal tip speed triggers fires when the predicted tip speed crosses \SI{0.4}{m/s}; and the neighboring $S_2$ triggers fires at its peak bend rate. The first trigger uses local information, whereas the latter two use communicated distal and neighboring information, respectively. Event-driven timing uses $t_{\mathrm{rel}}=\Pi_{\mathcal T}(t_e+\delta)$, with offset $\delta$ selected on 100 targets, and is evaluated on 400 disjoint targets.

\subsection{Constrained learning objective}
Policy search combines three hardware-based constraints: a \emph{snap set} of proximal preloads supported by the classifier; the independent sweep's \SIrange{0.45}{0.675}{s} \emph{timing set}; and a \emph{model-support set} favoring recorded-hold proximity and GRU-seed agreement. Unsupported candidates are penalized even when one model predicts a small miss.

At closest admissible contact, let $\ell$ be the cup-face error projected onto the target plane and $g$ its signed gap along the surface normal. Let $v_{\mathrm{pk}}$ and $t_{\mathrm{pk}}$ be the magnitude and time of peak tip speed, and $t_{\mathrm{ca}}$ the closest-approach time. With $\sigma(z)=(1+e^{-z})^{-1}$, the shaped contact ($A$), speed ($S$), and phase terms ($H$) are
\begin{align}
G&=\sigma\!\left(\frac{d_f-|g|}{\sigma_g}\right),\qquad
A=\frac{G}{2}\sum_{q\in\{a,b\}}
\exp\!\left(-\frac{\ell^2}{2\sigma_q^2}\right),\notag\\
S&=\sigma\!\left(\frac{v_{\mathrm{pk}}-v_{\min}^{\mathrm{sim}}}{v_s}\right),
\qquad
H=\exp\!\left[-\frac{(t_{\mathrm{ca}}-t_{\mathrm{pk}})^2}
{2\sigma_t^2}\right].
\label{eq:reward-terms}
\end{align}
The constants are $d_f=\SI{8}{cm}$, $\sigma_g=\SI{2}{cm}$, $(\sigma_a,\sigma_b)=(\SI{3}{cm},\SI{10}{cm})$, $v_{\min}^{\mathrm{sim}}=\SI{0.80}{m/s}$, $v_s=\SI{0.12}{m/s}$, and $\sigma_t=\SI{0.10}{s}$. The reduced simulation threshold compensates for the model's measured speed underprediction near the hardware attachment condition \SI{1}{m/s}. The complete rollout reward is
\begin{equation}
R=A S\left(\frac12+\frac12H\right) C_h
\min\!\left\{1,\frac{P_{\mathrm{snap}}(\rho_1,\rho_2)}{0.5}\right\}
-\lambda\bar\sigma_{\mathrm{tip}},
\label{eq:reward}
\end{equation}
where $C_h$ is the hold-support confidence defined above, $P_{\mathrm{snap}}$ is the hardware-fitted snap probability, $\bar\sigma_{\mathrm{tip}}$ is mean ensemble disagreement along the tip trajectory, and $\lambda=1.5\,\mathrm m^{-1}$. CMA-ES maximizes the Monte Carlo estimate of expected reward over the reachable target distribution.

The shaped reward guides optimization; final simulated attachment uses the event in Eq.~\eqref{eq:catch}. Its \SI{5}{cm} reporting tolerance is not a hard threshold in the learning objective. Table~\ref{tab:evidence} summarizes the evidence sets and their separation from the final evaluation.

\begin{table*}[t]
\caption{Evidence sets, their roles, and separation between development and evaluation.}
\vspace{-3mm}
\label{tab:evidence}
\centering
\scriptsize
\setlength{\tabcolsep}{2pt}
\renewcommand{\arraystretch}{0.88}
\begin{tabular}{@{}p{0.15\textwidth}p{0.06\textwidth}p{0.08\textwidth}p{0.22\textwidth}p{0.39\textwidth}@{}}
\toprule
Evidence & Domain & $n$ & Use & Separation \\
\midrule
Servo calibration & HW & 291 & Actuator limit & Independent recordings \\
Attachment trials & HW & 90 & Speed/angle condition & Run-grouped cross-validation \\
Delay sweep & HW & 33 & Timing/dynamics gate & Excluded from model fitting \\
Dynamics corpus & HW & 538 & Hold and snap models & Group split: 440 train, 78 validation, 20 sealed test \\
Policy targets & Sim. & 400 & Generalisation & Disjoint after 64-target selection \\
Release targets & Sim. & 400 & Timing ablation & Offsets fitted on 100 other targets \\
Architectures & Sim. & 5 seeds & Reward, attachment, efficiency & Common model, targets, bounds, optimiser, and budget \\
Deployment & HW & 18 & Retained attachment & Post-export; no retraining \\
\bottomrule
\end{tabular}
\vspace{-3mm}
\end{table*}

\section{Simulation Results}
\subsection{Protocol and task metric}
All results in this section are produced by the three-seed forward-model ensemble. Candidate policies start from model-predicted pre-release configurations. The same mapping converts the six policy parameters into nine servo goals in simulation and on hardware. The 400 final targets are unused for model fitting, policy optimization, model selection, or event-offset selection. For trial $i$, $\ell_i$ and $g_i$ denote the lateral error and normal gap at closest approach. Simulated attachment at lateral tolerance $r$ is defined by
\begin{equation}
a_i(r)=\mathbb I\!\left[\ell_i\le r,\ |g_i|\le d_f,\
v_{i,\mathrm{pk}}\ge v_{\min}^{\mathrm{sim}}\right].
\label{eq:catch}
\end{equation}
Equation~\eqref{eq:catch} requires lateral overlap, an admissible compression gap $d_f=\SI{8}{cm}$, and peak speed above $v_{\min}^{\mathrm{sim}}=\SI{0.80}{m/s}$. We report the full lateral-tolerance sweep because \SI{5}{cm} is not a physical suction-capture tolerance, and retained attachment is tested separately in Sec.~\ref{sec:hardware}.

\begin{figure*}[t]
\centering
\includegraphics[width=0.90\textwidth]{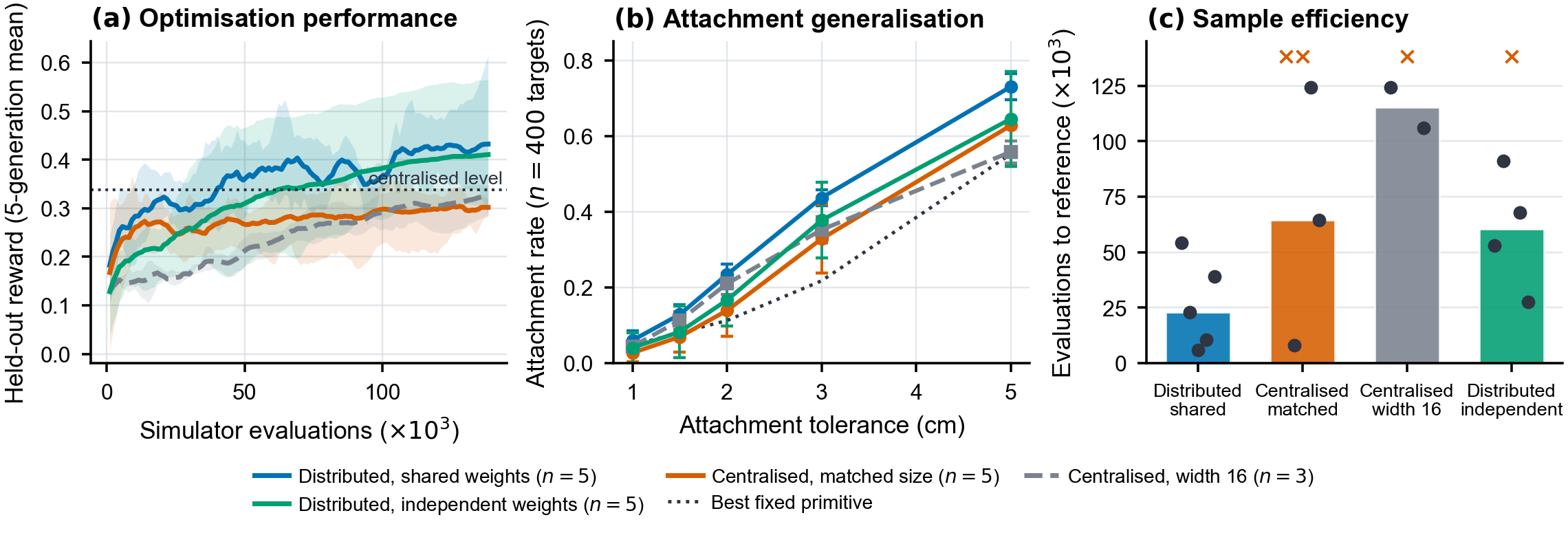}
\vspace{-5mm}
\caption{Architecture benchmark under a common simulation protocol. (a) Held-out reward versus evaluations; lines are means and bands span optimisation seeds. (b) Simulated attachment rate on 400 unseen targets versus geometric tolerance (mean $\pm$ standard deviation across seeds). (c) Evaluations required to reach the matched centralised policy's mean final reward; crosses mark seeds that do not reach it within budget.}
\label{fig:benchmark}
\vspace{-3mm}
\end{figure*}

The reward curves show the means and ranges in optimization seeds (Fig.~\ref{fig:benchmark}(a)); attachment summaries report the mean and standard deviations of the seeds. Release-rule rates use Wilson intervals over 400 targets and paired McNemar tests. Timing-reward intervals bootstrap target families to account for repeated starts. Architecture rewards use two-sided Mann--Whitney tests over seeds. Five seeds per principal architecture limit inference from nonsignificant attachment-rate differences.

\subsection{Generalization and allocation of the primitive}
On 400 unseen targets over a \SI{180}{\degree} bearing arc, the selected shared policy has closest-approach percentiles of \SIlist{16;38;86}{mm}, versus \SIlist{22;52;103}{mm} for the best of 48 fixed primitives. At the \SI{5}{cm} reporting tolerance, its simulated attachment rate is 0.74, versus 0.55 for the fixed primitive (Fig.~\ref{fig:timing}(c)). Every strike exceeds the calibrated speed threshold. Median miss varies from \SIrange{32}{42}{mm} across six \SI{30}{\degree} bearing sectors and more strongly with target distance.

The selected commands stay within the recorded hardware envelope: $\rho_1=2960$--3230, $\rho_2=2620$--2880, and $\rho_3=2660$--3130 ticks, with 0--50 ticks of tip slack. The distal preload correlates with the height of the target ($+0.78$) and the distance ($+0.69$), while the middle preload varies more moderately with distance ($+0.49$). S1 supplies the main targeting corrections: aim correlates $-0.93$ with height and release time $-0.73$ with distance. Release spans \SIrange{0.45}{0.67}{s}, with no evaluation target within a timing limit. Section identity thus routes common target features to distinct actions without an explicit division-of-labor reward.

In 62 simulated successes used for kinematic analysis, a curvature-weighted bend point moves distally in 93.5\% of cases (mean time--arc-length correlation 0.74). This is consistent with the sequential redistribution of elastic energy, but it does not establish a biological activation wave.

\subsection{Architecture and release information}
The shared policy obtains a higher final held-out reward than the matched centralized controller ($0.49\pm0.06$ versus $0.34\pm0.03$; two-sided Mann--Whitney $p=0.008$, five seeds each; Fig.~\ref{fig:benchmark}(a)). All five seeds of shared policies reach the centralized final-reward level after a median evaluation $2.30\times10^4$, 17\% of the budget (Fig.~\ref{fig:benchmark}(c)). Only three of five matched centralized seeds reach it (median $6.45\times10^4$). At \SI{5}{cm}, simulated attachment rates are $0.73\pm0.04$ and $0.63\pm0.10$, respectively (Fig.~\ref{fig:benchmark}(b)); with five seeds, this difference is not significant ($p=0.10$--0.15 across the reported attachment comparisons). The supported architecture claim thus is higher reward and greater sample efficiency, not statistical superiority in terminal simulated attachment rate.

The wider and unshared controls clarify this result. Only two of the three width-16 centralized seeds reach the common reward level, after a median evaluation $1.15\times10^5$, and their simulated attachment rate \SI{5}{cm} is $0.56\pm0.03$. Four of five unshared distributed seeds reach the level (median $6.05\times10^4$ evaluations) and obtain $0.65\pm0.13$. The best matched centralized seed reaches 0.81, so that the network can represent a strong solution; its optimization is less consistent under the common budget. The shared result supports a useful section-wise search bias, not an expressivity bound on centralized control.

\begin{figure*}[t]
\centering
\includegraphics[width=0.88\textwidth]{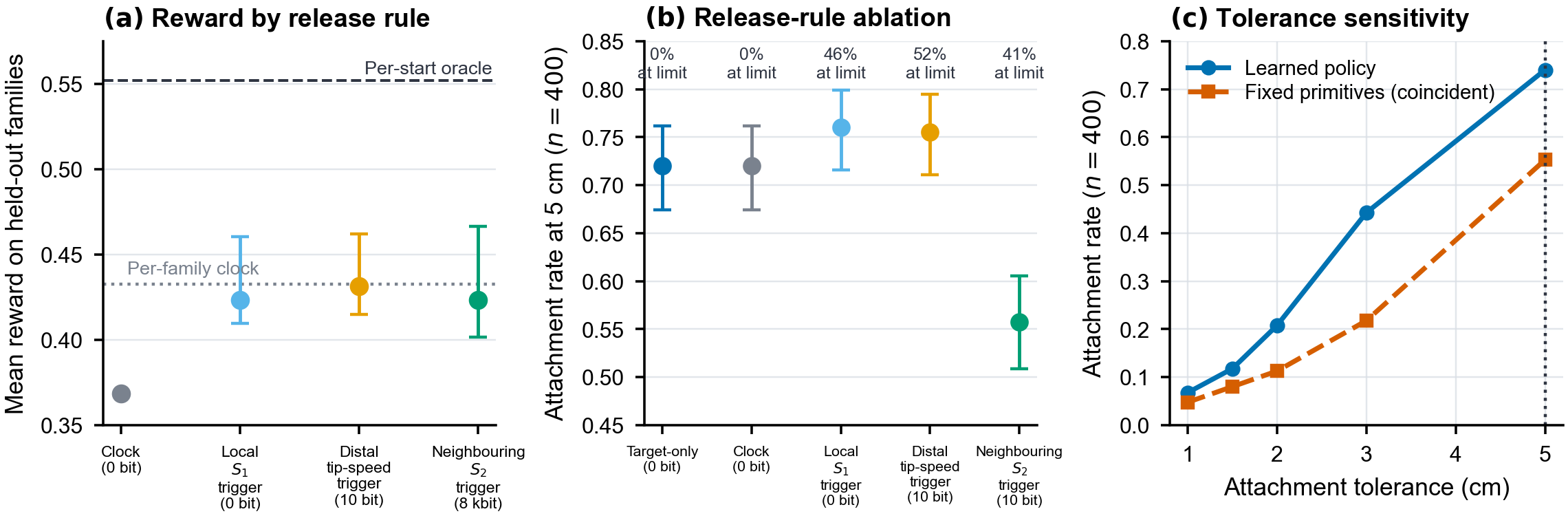}
\vspace{-3mm}
\caption{Release-rule ablation in simulation. (a) S1 timing reward; error bars are target-family bootstrap 95\% intervals. (b) Simulated attachment rate at the \SI{5}{cm} reporting tolerance ($n=400$); bars are Wilson 95\% intervals and annotations show the fraction projected to a validated timing bound. (c) Simulated attachment rate versus geometric tolerance for the learned policy and fixed primitives.}
\label{fig:timing}
\vspace{-6mm}
\end{figure*}

Event triggers improve the mean timing reward on the single clock but remain near the per-family clock and below the per-start oracle (Fig.~\ref{fig:timing}(a)). At \SI{5}{cm}, target-only and fixed-clock release each achieve 0.720 attachment (Fig.~\ref{fig:timing}(b)). The local $S_1$ trigger reaches 0.760 ($p=0.036$, paired McNemar); distal tip speed reaches 0.755 ($p=0.103$); and the neighboring $S_2$ drops to 0.558 ($p<10^{-10}$). These triggers project 46\%, 52\%, and 41\% of releases, respectively, to a timing bound. Local geometry offers a small gain, but the tested inter-section events establish no advantage over target-only control.

In 133 held-base hardware trials, the S1 servo-position residual shows a 12-tick trigger transient and otherwise stays near 3 ticks, indicating that the non-backdrivable gearbox suppresses the mechanical wave in motor position. A causal base-deformation detector fires within \SI{100}{ms} of the event in 70\% of the trials, versus 55\% for the best clock, 64\% for the servo residual and 65\% with the neighbor state. Body geometry thus provides useful local timing information.

Perturbing the predicted pre-release position and orientation at 0.5, 1, 1.5 and 2 times the leave-one-session-out error scale increases the median simulated miss from \SI{38}{mm} (\SI{5.2}{\percent} of $L_{\mathrm{arm}}$) to \SIlist{43;53;65;78}{mm} (\SIlist{5.9;7.3;8.9;10.7}{\percent}), respectively. Perturbations that match the measured validation-error scale therefore increase the median miss by 39\%, while twice that scale more than doubles it. Mean predicted peak speed decreases from 1.18 to 0.98 model units across this range. The physical experiment therefore tests the complete loading-and-release pipeline rather than reproducing a simulated attachment probability.

Failures can arise from insufficient snap energy, mistimed release, or a fast snap that misses the target or lacks cup compression. The selected simulated strikes all exceed the speed threshold, but attachment remains strongly tolerance-dependent (Fig.~\ref{fig:timing}(c)), identifying aim as the dominant simulated limitation. Hardware additionally introduces contact and seal failures that are absent from the model.

\section{Hardware Demonstration}\label{sec:hardware}
\begin{figure}[t]
\centering
\includegraphics[width=\linewidth]{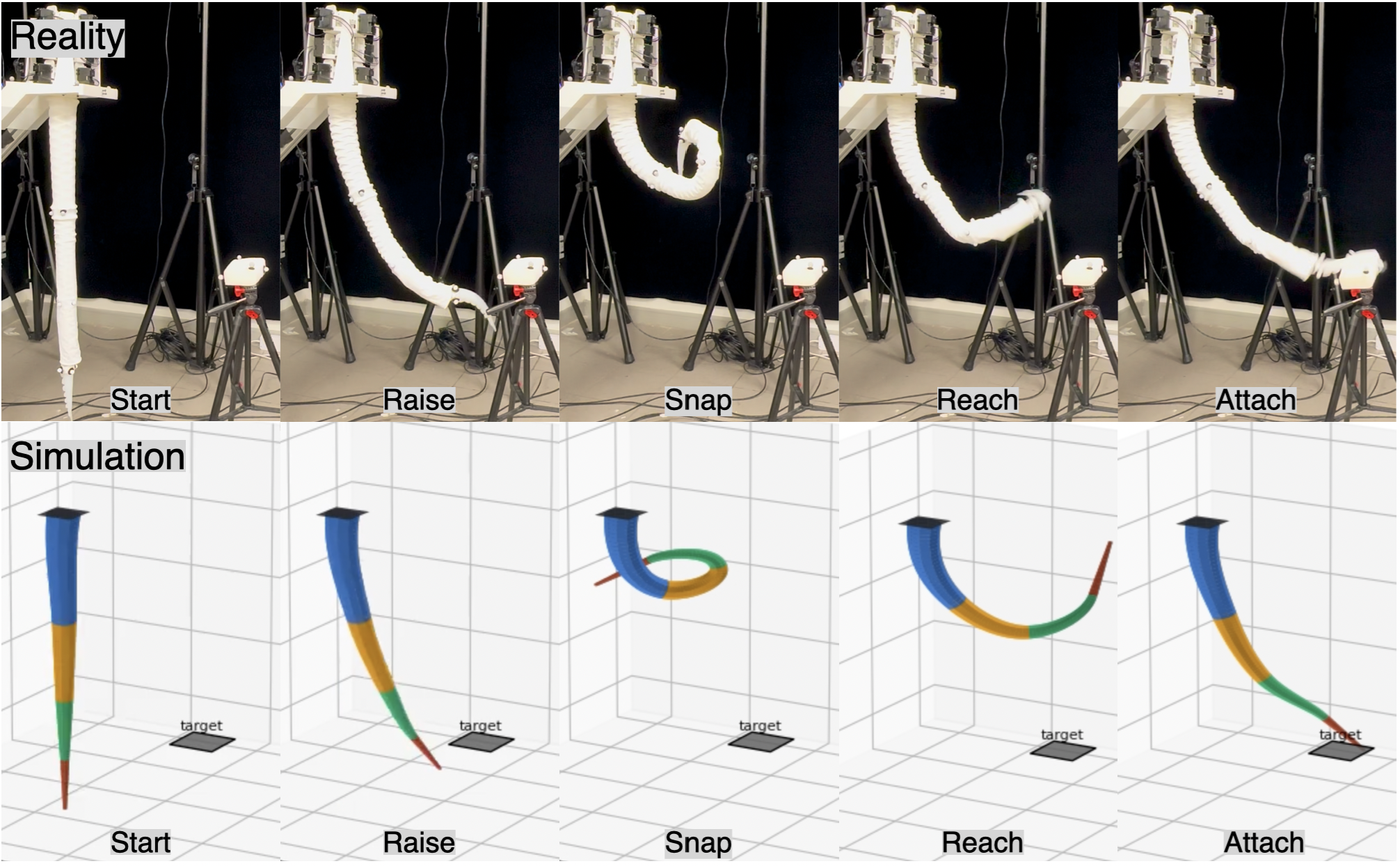}
\vspace{-2mm}
\caption{Representative successful execution of the frozen learned policy in hardware (top) and simulation (bottom): hanging arm, preload, elastic snap, target reach, and passive attachment.}
\label{fig:transfer}
\vspace{-1mm}
\end{figure}

\begin{table}[t]
\caption{Hardware evidence chain for the learnt policy.}
\vspace{-3mm}
\label{tab:hardware}
\centering
\footnotesize
\begin{tabular}{@{}>{\raggedright\arraybackslash}p{0.26\linewidth}
>{\raggedright\arraybackslash}p{0.40\linewidth}
>{\raggedright\arraybackslash}p{0.24\linewidth}@{}}
\toprule
Evidence & Criterion & Result \\
\midrule
Preload reached & S2 bend increase $\geq\SI{110}{\degree}$ & 18/18 \\
Fast snap produced & Peak tip speed $\geq\SI{1}{m/s}$ & 17/18 \\
Snap-speed distribution & Median [IQR] & \SI{1.33}{m/s} [1.19--1.44] \\
Object retrieved & Detached and carried & 9/18 \\
Spatial replication & Placements with $\geq1$ retrieval & 5/6 \\
Online correction & State feedback during snap & None \\
\bottomrule
\end{tabular}
\vspace{-2mm}
\end{table}

We evaluated the end-to-end deployment of the learnt shared policy in the physical arm (Fig.~\ref{fig:transfer}). Eighteen open-loop executions were completed at six spatially distinct placements spanning commanded target distances of \SIrange{0.62}{0.69}{m} and a \SI{91}{\degree} bearing range. The free-standing target block measured approximately \SI{10}{cm}$\times$\SI{6}{cm}$\times$\SI{3}{cm}. Motion capture provided the target position before each execution, and no retraining of the state feedback policy or adjustment of target-specific parameters was used during movement.

The controller reliably executed the mechanical primitive (Table~\ref{tab:hardware}). All 18 trials reached the prescribed preloaded configuration, and 17 exceeded the hardware-derived snap threshold of \SI{1}{m/s}. Based on the \SI{240}{Hz} motion-capture traces, the peak tip speed had a median of \SI{1.33}{m/s}, an IQR of \SIrange{1.19}{1.44}{m/s}, and a maximum of \SI{1.61}{m/s}. The system successfully detached and retained the object in nine trials, achieving retrieval at five of the six tested placements. The failure in the remaining placement was attributed to the geometric asymmetry of the arm.

Table~\ref{tab:hardware} separates execution of the mechanical primitive from completion of the retrieval task. All 18 trials reached the prescribed preload, 17 produced a fast snap, and nine retrieved the object across five of six placements. This gap shows that sufficient snap speed alone does not guarantee retrieval and motivates further investigation of target alignment, contact geometry, and suction sealing. Nearly every execution generated sufficient snap speed, but only nine formed a retained attachment. The remaining limitations arise downstream and involve target alignment, contact geometry, and suction seal strength. These factors could be addressed through real-time sensor feedback and improved suction cup design, both of which are beyond the scope of this paper. Therefore, the experiments demonstrate the transfer of the complete loading-and-release pipeline and successful physical task execution across distinct target commands.

\section{Discussion}
The results support a division of responsibility in which learned control prepares the body and passive mechanics execute the rapid movement. Octopus biology motivates the section-wise organizational prior, without implying reproduction of the octopus nervous system. The architecture comparisons support shared policy weights as an effective parameterization for optimization, but do not establish that local processing is necessary. The hardware-gated simulator's role is limited. The simulator’s approximately \SI{60}{mm} hold error, reduced predicted speed under state perturbations, finite validation horizon, and omission of contact and suction-seal dynamics limit quantitative sim-to-real claims. Simulation therefore supports comparisons only within the validated operating region. The hardware evaluation remains exploratory because it includes only a limited number of trials across six placements. Future studies should evaluate a broader range of placements and targets, apply fixed retention and transport criteria, and classify failures as energetic, temporal, geometric, or seal-related. In addition, experiments with additional arm sections while keeping the shared policy weights fixed would provide a test of scalability.

\section{Conclusion}
This work demonstrates that actuator speed does not need to determine the dynamic limit of a compliant arm. By selecting target-dependent preload and release conditions, controlled elastic release increased the tip speed to at least \SI{108}{\percent} above direct tendon-driven release. Although evaluation remains limited, the results show that a compact distributed controller can harness compliant-body mechanics as a source of rapid motion, extending a slow-actuated soft arm from quasi-static positioning to dynamic reach-and-attach. This demonstrates how embodied intelligence can expand the capabilities of robotic tasks under fixed actuator constraints as a learned policy selects the conditions for energy storage and release, while the compliant body supplies the rapid motion needed for goal-directed reach-and-attach.

\section*{Acknowledgment}
Claude and ChatGPT were used for limited assistance with language, code implementation, and visualization in this paper. The authors have verified all content.

\bibliographystyle{IEEEtran}
\bibliography{refs}

@article{pfeifer2007self,
  title   = {Self-organization, embodiment, and biologically inspired robotics},
  author  = {Pfeifer, Rolf and Lungarella, Max and Iida, Fumiya},
  journal = {Science},
  volume  = {318},
  number  = {5853},
  pages   = {1088--1093},
  year    = {2007},
  doi     = {10.1126/science.1145803}
}

@article{sitti2021physical,
  title   = {Physical intelligence as a new paradigm},
  author  = {Sitti, Metin},
  journal = {Extreme Mechanics Letters},
  volume  = {46},
  pages   = {101340},
  year    = {2021},
  doi     = {10.1016/j.eml.2021.101340}
}

@article{mengaldo2022concise,
  title   = {A concise guide to modelling the physics of embodied intelligence in soft robotics},
  author  = {Mengaldo, Gianmarco and Renda, Federico and Brunton, Steven L. and B{\"a}cher, Moritz and Calisti, Marcello and Duriez, Christian and Chirikjian, Gregory S. and Laschi, Cecilia},
  journal = {Nature Reviews Physics},
  volume  = {4},
  number  = {9},
  pages   = {595--610},
  year    = {2022},
  doi     = {10.1038/s42254-022-00481-z}
}

@article{milana2025physical,
  title   = {Physical control: A new avenue to achieve intelligence in soft robotics},
  author  = {Milana, Edoardo and Della Santina, Cosimo and Gorissen, Benjamin and Rothemund, Philipp},
  journal = {Science Robotics},
  volume  = {10},
  number  = {102},
  pages   = {eadw7660},
  year    = {2025},
  doi     = {10.1126/scirobotics.adw7660}
}

@article{ilton2018principles,
  author  = {Ilton, Mark and
             Bhamla, M. Saad and
             Ma, Xiaotian and
             Cox, Suzanne M. and
             Fitchett, Leah L. and
             Kim, Yongjin and
             Koh, Je-Sung and
             Krishnamurthy, Deepak and
             Kuo, Chi-Yun and
             Temel, Fatma Zeynep and
             Crosby, Alfred J. and
             Prakash, Manu and
             Sutton, Gregory P. and
             Wood, Robert J. and
             Azizi, Emanuel and
             Bergbreiter, Sarah and
             Patek, S. N.},
  title   = {The principles of cascading power limits in small,
             fast biological and engineered systems},
  journal = {Science},
  volume  = {360},
  number  = {6387},
  pages   = {eaao1082},
  year    = {2018},
  doi     = {10.1126/science.aao1082}
}

@article{pal2020elastic,
  title   = {Elastic Energy Storage Enables Rapid and Programmable Actuation in Soft Machines},
  author  = {Pal, Aniket and Goswami, Debkalpa and Martinez, Ramses V.},
  journal = {Advanced Functional Materials},
  volume  = {30},
  number  = {1},
  pages   = {1906603},
  year    = {2020},
  doi     = {10.1002/adfm.201906603}
}

@article{gorissen2020inflatable,
  author  = {Gorissen, Benjamin and
             Melancon, David and
             Vasios, Nikolaos and
             Torbati, Mehdi and
             Bertoldi, Katia},
  title   = {Inflatable soft jumper inspired by shell snapping},
  journal = {Science Robotics},
  volume  = {5},
  number  = {42},
  pages   = {eabb1967},
  year    = {2020},
  doi     = {10.1126/scirobotics.abb1967}
}

@article{sumbre2001control,
  title={Control of octopus arm extension by a peripheral motor program},
  author={Sumbre, German and Gutfreund, Yoram and Fiorito, Graziano and Flash, Tamar and Hochner, Binyamin},
  journal={Science}, volume={293}, number={5536}, pages={1845--1848}, year={2001}, doi={10.1126/science.1060976}}

@article{hochner2023embodied,
  title={Embodied mechanisms of motor control in the octopus},
  author={Hochner, Binyamin and Zullo, Letizia and Shomrat, Tal and Levy, Guy and Nesher, Nir},
  journal={Current Biology}, volume={33}, number={20}, pages={R1119--R1125}, year={2023}, doi={10.1016/j.cub.2023.09.008}}

@article{kier2002structure,
  title={The structure and adhesive mechanism of octopus suckers},
  author={Kier, William M. and Smith, Andrew M.},
  journal={Integrative and Comparative Biology}, volume={42}, number={6}, pages={1146--1153}, year={2002}, doi={10.1093/icb/42.6.1146}}

@article{mazzolai2019octopus,
  title={Octopus-inspired soft arm with suction cups for enhanced grasping tasks in confined environments},
  author={Mazzolai, Barbara and Mondini, Alessio and Tramacere, Francesca and Riccomi, Gianluca and Sadeghi, Ali and Giordano, Goffredo and Del Dottore, Emanuela and Scaccia, Massimiliano and Zampato, Massimo and Carminati, Stefano},
  journal={Advanced Intelligent Systems}, volume={1}, number={6}, pages={1900041}, year={2019}, doi={10.1002/aisy.201900041}}

@inproceedings{hou2026quantitative,
  title={A Quantitative Comparison of Centralised and Distributed Reinforcement Learning-Based Control for Soft Robotic Arms},
  author={Hou, Linxin and Wu, Qirui and Qin, Zhihang and Banerjee, Neil and Guo, Yongxin and Laschi, Cecilia},
  booktitle={2026 IEEE 9th International Conference on Soft Robotics (RoboSoft)},
  address={Kanazawa, Japan}, year={2026}, pages={475--480}, doi={10.1109/RoboSoft67810.2026.11522945}}

@inproceedings{hou2026octopus,
  title={Octopus-inspired Distributed Control for Soft Robotic Arms: A Graph Neural Network-Based Attention Policy with Environmental Interaction},
  author={Hou, Linxin and Wu, Qirui and Qin, Zhihang and Guo, Yongxin and Laschi, Cecilia},
  booktitle={Proc. IEEE/RSJ Int. Conf. Intelligent Robots and Systems (IROS)},
  year={2026},
  note={To appear}}

@article{margheri2012soft,
  title   = {Soft robotic arm inspired by the octopus: {I}. {F}rom biological functions to artificial requirements},
  author  = {Margheri, Laura and Laschi, Cecilia and Mazzolai, Barbara},
  journal = {Bioinspiration \& Biomimetics},
  volume  = {7},
  number  = {2},
  pages   = {025004},
  year    = {2012},
  doi     = {10.1088/1748-3182/7/2/025004}
}

@article{feng2025impulsive,
  title   = {Impulsive actuation for soft robots},
  author  = {Feng, Ruoyu and He, Yiming and Feng, Siyuan and Li, Shuguang},
  journal = {npj Robotics},
  volume  = {3},
  pages   = {27},
  year    = {2025},
  doi     = {10.1038/s44182-025-00045-0}
}

@article{liu2026underwater,
  title   = {Underwater soft arm grasping with simplified control using octopus-inspired bending propagation},
  author  = {Liu, Jiaqi and Zhu, Zhichao and Wen, Li},
  journal = {npj Robotics},
  volume  = {4},
  pages   = {2},
  year    = {2026},
  doi     = {10.1038/s44182-025-00066-9}
}

@article{falotico2025learning,
  title={Learning controllers for continuum soft manipulators: Impact of modeling and looming challenges},
  author={Falotico, Egidio and Donato, Enrico and Alessi, Carlo and Setti, Elisa and Nazeer, Muhammad Sunny and Agabiti, Camilla and Caradonna, Daniele and Bianchi, Diego and Piqu{\'e}, Francesco and Ansari, Yasmin Tauqeer and others},
  journal={Advanced Intelligent Systems},
  volume={7},
  number={2},
  pages={2400344},
  year={2025},
  publisher={Wiley Online Library}
}

@article{TRO_nazeer2024,
  title={Rl-based adaptive controller for high precision reaching in a soft robot arm},
  author={Nazeer, Muhammad Sunny and Laschi, Cecilia and Falotico, Egidio},
  journal={IEEE Transactions on Robotics},
  volume={40},
  pages={2498--2512},
  year={2024},
  publisher={IEEE}
}

@article{renda2018discrete,
  title   = {Discrete {Cosserat} Approach for Multisection Soft Manipulator Dynamics},
  author  = {Renda, Federico and Boyer, Fr{\'e}d{\'e}ric and Dias, Jorge and Seneviratne, Lakmal},
  journal = {IEEE Transactions on Robotics},
  volume  = {34},
  number  = {6},
  pages   = {1518--1533},
  year    = {2018},
  doi     = {10.1109/TRO.2018.2868815}
}

@article{hansen2001completely,
  title   = {Completely Derandomized Self-Adaptation in Evolution Strategies},
  author  = {Hansen, Nikolaus and Ostermeier, Andreas},
  journal = {Evolutionary Computation},
  volume  = {9},
  number  = {2},
  pages   = {159--195},
  year    = {2001},
  doi     = {10.1162/106365601750190398}
}

@inproceedings{wang2018nervenet,
  title     = {{NerveNet}: Learning Structured Policy with Graph Neural Networks},
  author    = {Wang, Tingwu and Liao, Renjie and Ba, Jimmy and Fidler, Sanja},
  booktitle = {International Conference on Learning Representations},
  year      = {2018},
  url       = {https://openreview.net/forum?id=S1sqHMZCb}
}

@inproceedings{gupta2022metamorph,
  title     = {{MetaMorph}: Learning Universal Controllers with Transformers},
  author    = {Gupta, Agrim and Fan, Linxi and Ganguli, Surya and Fei-Fei, Li},
  booktitle = {International Conference on Learning Representations},
  year      = {2022},
  url       = {https://openreview.net/forum?id=Opmqtk\_GvYL}
}

@inproceedings{berdica2024reinforcement,
  title     = {Reinforcement Learning Controllers for Soft Robots Using Learned Environments},
  author    = {Berdica, Uljad and Jackson, Matthew and Veronese, Niccol{\`o} Enrico and Foerster, Jakob and Maiolino, Perla},
  booktitle = {2024 IEEE 7th International Conference on Soft Robotics (RoboSoft)},
  pages     = {933--939},
  year      = {2024},
  doi       = {10.1109/RoboSoft60065.2024.10522003}
}

@article{tang2026general,
  title   = {A general soft robotic controller inspired by neuronal structural and plastic synapses that adapts to diverse arms, tasks, and perturbations},
  author  = {Tang, Zhiqiang and Tian, Liying and Xin, Wenci and Wang, Qianqian and Rus, Daniela and Laschi, Cecilia},
  journal = {Science Advances},
  volume  = {12},
  number  = {2},
  pages   = {eaea3712},
  year    = {2026},
  doi     = {10.1126/sciadv.aea3712}
}

@article{haggerty2023inertial,
  title   = {Control of soft robots with inertial dynamics},
  author  = {Haggerty, David A. and Banks, Michael J. and Kamenar, Ervin and Cao, Alan B. and Curtis, Patrick C. and Mezi{\'c}, Igor and Hawkes, Elliot W.},
  journal = {Science Robotics},
  volume  = {8},
  number  = {81},
  pages   = {eadd6864},
  year    = {2023},
  doi     = {10.1126/scirobotics.add6864}
}

@article{bianchi2024softoss,
  title   = {{SofToss}: Learning to Throw Objects With a Soft Robot},
  author  = {Bianchi, Diego and Antonelli, Michele Gabrio and Laschi, Cecilia and Sabatini, Angelo Maria and Falotico, Egidio},
  journal = {IEEE Robotics \& Automation Magazine},
  volume  = {31},
  number  = {4},
  pages   = {113--123},
  year    = {2024},
  doi     = {10.1109/MRA.2023.3310865}
}
\end{document}